\documentclass[11pt]{article}

\usepackage[]{acl}

\usepackage{times}
\usepackage{latexsym}
\usepackage[T1]{fontenc}
\usepackage[utf8]{inputenc}
\usepackage{microtype}
\usepackage{inconsolata}
\usepackage{graphicx}
\usepackage{booktabs}
\usepackage{amsmath}
\usepackage{subcaption}
\usepackage{xspace}
\usepackage{epigraph}
\newcommand{\corpus}{\texttt{bb24.train}\xspace}
\newcommand{\deberta}{DeBERTa-v3\xspace}
\newcommand{\vpswap}{VP-Swap\xspace}
\newcommand{\sammod}{75k-SAM\xspace}       
\newcommand{\basemod}{75k\xspace}          
\newcommand{\extmod}{75k-SAM-ext\xspace}   
\newcommand{\comps}{COMPS\xspace}
\newcommand{\glue}{GLUE\xspace}
\newcommand{\blimp}{BLiMP\xspace}
\newcommand{\ewok}{EWoK\xspace}

\title{Augustinian BabyLM: \\What Ostensive Definition Can and Cannot Teach a Small Language Model}

\author{Lisa Bylinina \\
  Institute for Language Sciences \\
  Utrecht University \\
  \texttt{e.g.bylinina@uu.nl} \\}

\begin{document}
\maketitle

\begin{abstract}

A language model normally begins training with random word embeddings:
whatever \emph{banana} means must be learned from training corpora. I implement St. Augustine's picture of word learning, meaning by ostension,
for a small masked language model (DeBERTa) trained on $\approx$10M words: before training, visually grounded tokens receive embeddings derived from
the image regions they label; other tokens start random.

Visual initialization leaves a measurable imprint that lasts until the end of training. At the same time, the effect remains invisible under most BabyLM benchmarks, which probe abstract grammatical knowledge: visual initialization does not affect performance there. The only zero-shot exception is object-property knowledge (COMPS, \citealt{misra2023comps}), where seeding helps in every configuration. To follow up on this result, I build a corpus-tailored version of the Visual-Property Swap benchmark
\citep{egobabyvlm}, which tests color, material, size, and shape knowledge, with per-item training frequency and seeded status. Here, vision-seeded models prove to have a persistent, seed-replicated advantage. The benchmark cannot, however, attribute this advantage to the seeded words themselves, due to a seededness-frequency confound.

Function words and abstract vocabulary also receive strong visual seeds and retain them throughout training, and the training objective draws on them: held-out mask-prediction loss falls for these words in every seed. However, no benchmark I run registers this. What evaluation would pick this up remains an open question.

\end{abstract}

\section{Introduction}

\epigraph{%
When they (my elders) named some object \ldots\ I grasped that the thing
was called by the sound they uttered.%
}{\textit{Confessions} I.8.13 \citep{augustine_confessions}}

Augustine's account of how he learned to speak is 
probably the oldest theory of lexical acquisition we have: elders point at things and say their names, and the child, seeing what is pointed at, binds sound to object.
\citet{wittgenstein_pi} opened the \emph{Philosophical Investigations}
by arguing against this view: ostension cannot be the basis of word learning, because pointing presupposes that the learner already parses the scene into objects and properties, and in that sense already has the space of lexical meanings ready. On the other hand, core-knowledge research \citep{spelke2007core} demonstrates exactly this kind of prior in newborns, with cognitive systems for objects, number, agents, and space. So, the matter is not settled.

\begin{figure*}[t]
  \centering
  \includegraphics[width=\textwidth]{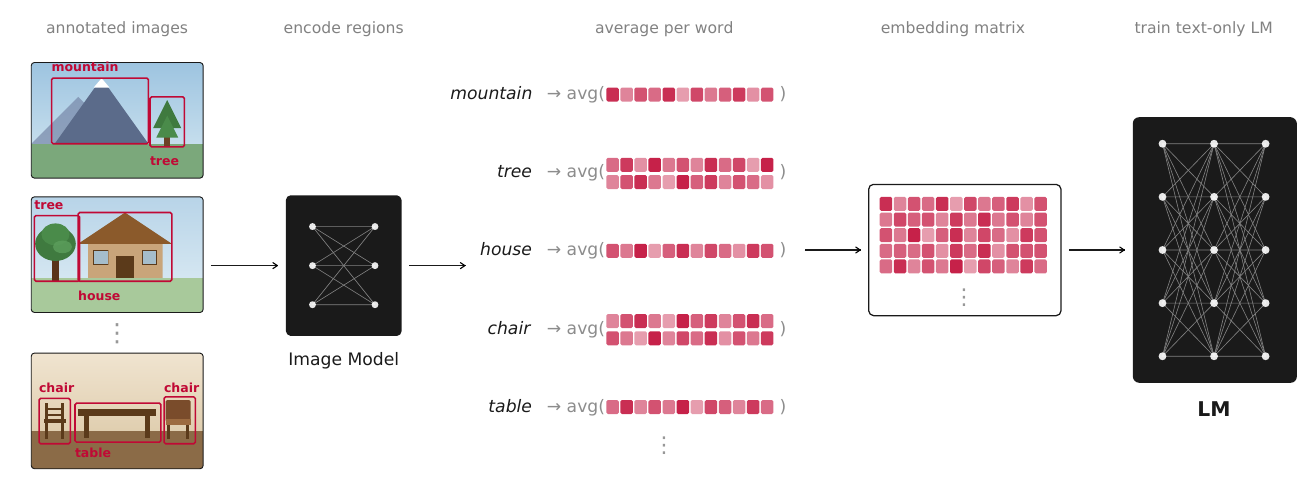}
  \caption{The vision-seeding pipeline. For each word that labels regions
  in the visual grounding data, I encode those regions with a frozen
  vision model, average the region features into a single vector, and
  write it (projected and scaled) into the corresponding row of the
  language model's input embedding matrix; words with no image support
  keep their random initialization. The language model is then trained on
  text only.}
  \label{fig:pipeline}
\end{figure*}

I look for effects of visual demonstration in language learning using computational methods, in particular, in language models. A typical  text-only language model is not equipped for ostension: its word
embeddings begin as random vectors, and their meanings must be discovered
from distribution alone. For comparison, I supply the visual demonstration: before any text is seen, words that label regions in visual data are initialized from a vision encoder's features over those regions (Fig.~\ref{fig:pipeline}); the model undergoes text-only training from that point on. The BabyLM strict-small budget, roughly a child's language input without the embodied component, is where this prior is expected to  
matter \citep[cf.][]{zhuang2024grounding}. I quantify its effect
against a matched text-only baseline.\footnote{Code, models, and my version of the VP-Swap benchmark are
available at \url{https://github.com/bylinina/augustinian_babylm}. All
trained models, the per-region visual features, and the per-token
embedding tables are released under the \texttt{augustinian-babylm}
organization on the Hugging Face Hub:
\url{https://huggingface.co/augustinian-babylm}.}

I show that visual initialization has a measurable effect: its imprint on the embeddings lasts to the end of training, and the training objective itself becomes easier for the words that received visual grounding. This effect is, however, difficult to observe through the standard BabyLM benchmarks: barely any of them probe the aspects of meaning for which visual information is relevant. The exception is COMPS \citep{misra2023comps}, which tests object-property knowledge and improves in all configurations.

Following this observation, I build my version of \vpswap \citep{egobabyvlm}, a visual-property minimal-pair benchmark. There the advantage is clear: it is persistent and replicated across seeds. The benchmark turns out to be unable to show that the advantage is specific to visually grounded words -- the grounded words tend to be the frequent ones, leading to a seededness-frequency confound.

Abstract and function words complicate the picture: their visual seeds are also strong and well retained, and the objective draws on them, but nothing I measure rewards it.

\paragraph{Contributions.}
(1)~An effect of vision-seeded token embedding initialization for a small language model under the 10M-word training budget, replicated across configurations and random seeds and corresponding to a consistent object-property gain according to the \comps benchmark.
(2)~An adjusted version of \vpswap: a corpus-tailored visual-property benchmark with per-item training-frequency and seeded-status metadata.
(3)~A synthetic-grounding experiment designed as a causal test; a pair-level analysis shows it is uninformative, due to a frequency confound.
(4)~A mechanistic account of how the effect
survives the embeddings' departure from their visual anchors.
(5)~A scope result: for function words the visual seeds are strong, retained, and relevant for the learning objective, yet the effect is not detected by existing benchmarks.

\section{Related Work}

\paragraph{Multimodal learning under the BabyLM budget.}
Whether visual input helps models learn language in a more human-like way has been a recurring BabyLM question in the (now abandoned) multimodal track 
\citep{choshen2024findings}. Results have been modest so far: submissions have mostly
not beaten the baselines
\citep{choshen2024findings,charpentier2025findings}, and multimodal training has
sometimes harmed language-only performance, under the catastrophic forgetting effect when a text model is
fine-tuned on linguistically deficient caption data \citep{amariucai2023acquiring,klerings2024modular}.
Even approaches that beat the baselines inherit the problem:
\citet{takmaz2025merging} report 
models that beat prior baselines but fall behind on
grammar-focused text-only benchmarks, motivating inference-time model merging with a text-only model to recover the lost ability. My approach addresses the same issue in a different way: visual
information is supplied only once, as the initialization of a subset of
embedding rows, with no continued caption signal and no change
to the architecture or training objective. The model is a regular text-only masked language model.

\paragraph{Grounded word representations.}
The idea that word meaning has a perceptual component with measurable consequences for representation has a long history, from multimodal distributional semantics \citep{bruni2014multimodal} to the finding that grounded and text-only models decode brain activity differently for concrete versus abstract nouns \citep{anderson2017visually}. I reuse this
insight in a different setup: features from a
frozen vision encoder \citep{dinov3,ibot,kirillov2023segment} supply the starting embeddings for the words that name image regions. Unlike approaches that inject grounding through auxiliary objectives or cross-modal attention, my setting does not affect the model's parameters beyond the initial embedding values.

\paragraph{What targeted evaluations measure.}
The benchmarks that dominate small-model evaluation are mostly probes of abstract structural competence \citep{warstadt2020blimp}. Visual grounding is hardly expected to have a systematic effect here -- instead, it should be relevant for aspects of lexical semantics: knowledge of what words denote and what their referents are like. This is also linguistically relevant knowledge, since it governs
how words are used. So, an
intervention that affects meaning will surface only on evaluations that address word meaning.
Two such probes are important for me: \comps \citep{misra2023comps}, which tests knowledge of object properties, and \vpswap, a visual-property
minimal-pair benchmark I adapt from \citet{egobabyvlm}. Their version tested a different intervention and found no effect. I rebuild the benchmark from my corpus so that each item records how often the model saw each word and whether that word was visually seeded. This information was used to try and trace the effect to the specific words involved in the intervention. In \S\ref{sec:vpswap}, I discuss the complications to this setup.

\section{The Augustinian Setup}
\label{sec:setup}

My setup implements ostension as a single change to how training begins (Figure~\ref{fig:pipeline}). A language model's input embedding matrix has one row per vocabulary item, and normally every row
starts with random numbers. I replace this noise for some rows with a visual summary of the corresponding word: for each word that appears labeling regions of images in visual grounding data, I collect those regions, pass them through a frozen vision encoder, and average the resulting features into a single vector that becomes the word's initial embedding. Words that never label an image region keep their random start. Nothing else about
the model or its training changes: the architecture, the objective, and the text data belong to the regular text-only masked language model training setup. In this way, any difference I observe between a visually seeded model and the baseline is
attributable to this intervention. The rest of this section makes the two halves of
the procedure precise: how a word's (more specifically, token's) visual embedding is computed
(\S\ref{sec:region-features}), and which words can receive one at all
(\S\ref{sec:coverage}).

\subsection{Region Features and Seeding}
\label{sec:region-features}
For visual grounding I rely on a combination of four datasets that pair words with image regions:
Flickr30k Entities \citep{plummer2015flickr30k}, RefCOCO+
\citep{kazemzadeh2014referitgame}, RefCOCOg \citep{mao2016refcocog}, and
THINGS \citep{hebart2019things}, together 563k region annotations.
For a target word, I take every region it labels, crop the region, and
encode it with a frozen vision model; the region feature is the mean of
the encoder's patch embeddings inside the bounding box, and the word's
visual embedding is the average of these region features over all its
occurrences (a word that tokenizes into several subwords contributes to
each of the tokens). I report three encoders (DINOv3 \citealt{dinov3}, iBOT
ViT-B/16 \citealt{ibot}, and SAM ViT-B \citealt{kirillov2023segment}), which
differ in training objective but are used identically here. Importantly, all three encoders are trained on images alone with no
text or caption supervision, unlike vision--language encoders such as CLIP
\citep{radford2021clip} whose features are shaped by language: DINOv3 and iBOT are trained by self-distillation on unlabeled images, SAM is trained on image-mask pairs. The visual seeds therefore carry a
purely perceptual prior, uncontaminated by linguistic signal, which also means that the grounding adds no words to the model's language budget. 
All three encoders produce 768-dimensional features, matching \deberta-base's embedding size,
so no projection is needed; I only rescale each seeded vector (mean-centering,
$L_2$-normalizing, and scaling to the standard deviation of the model's
random initializer) so that seeded and unseeded rows are statistically comparable at the start of training.  Tokens whose word never labels a region are left at their random initialization. Only input embeddings are seeded; all other parameters,
and the tied output embeddings, start as usual.
 
\subsection{Coverage: What Gets Seeded}
\label{sec:coverage}
\begin{figure}[t]
  \centering
  \includegraphics[width=\columnwidth]{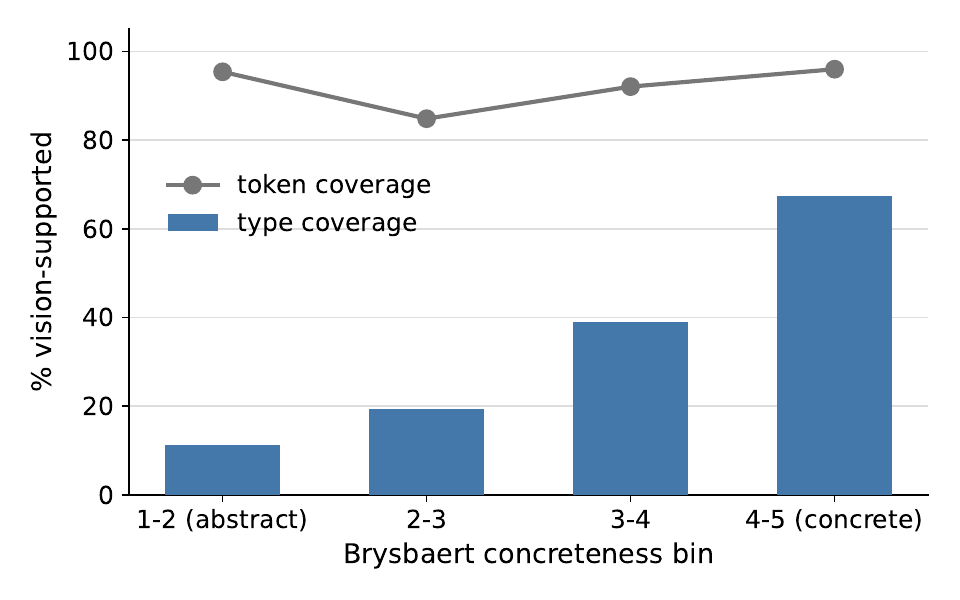}
  \caption{Seedability rises with concreteness. Word types binned by
  their Brysbaert concreteness rating; the height of each bar is the
  fraction of that bin appearing in the visual grounding data. Concrete
  words are far more likely to be seeded.
  }
  \label{fig:coverage}
\end{figure}

Because only words that label image regions can be seeded, the intervention reaches a subset of the vocabulary. Of the $\sim$163k word types in the
corpus, about $10\%$ ever appear in the grounding data, but these are common words, covering $85\%$ of text. Subword tokenization raises type coverage by sharing subwords: the seedable fraction of the vocabulary is $37.7\%$ at 50k merges, $29.1\%$ at 75k, and $23.8\%$ at 100k -- larger vocabularies  leave more rare whole-word rows un-seeded. Token
coverage stays near $87\%$ across vocabulary sizes. Seedable words skew strongly concrete (seededness correlates with concreteness norms at $r=0.46$;
Figure~\ref{fig:coverage}); predictably, verbs, discourse and abstract terms, and temporal expressions do not get strong visual support.

\subsection{Models and Training}
I train \deberta-base masked language models on \corpus, the
$\sim$9.9M-word corpus of \citet{edman2024babylm}, which mixes
LLM-generated paraphrase data with portions of the official BabyLM corpus (not the official 2026 strict-small distribution).
Crossing three BPE vocabulary sizes (50k/75k/100k) with the three image encoders, plus a random-initialized baseline per vocabulary, gives three
baselines and nine vision-init models, each trained for 10 epochs under a
fixed recipe (Appendix~\ref{app:hparams}) with checkpoints saved by
word count. Runs are single-seed except the headline comparison
(\sammod{} vs.\ \basemod) and the synthetic-extension model \extmod, which
I repeat with three seeds to address variance.

\subsection{Released Artifacts} 

All twelve models are public. Baselines are
named \texttt{deberta-base-}\textit{V} and vision-init models
\texttt{deberta-base-}\textit{V}\texttt{-}\textit{E}, for vocabulary
\textit{V} $\in$ \{50k, 75k, 100k\} and encoder \textit{E} $\in$ \{sam,
dinov3, ibot\}; three-seed replicates add a \texttt{-s1}, \texttt{-s2} or
\texttt{-s3} suffix. Intermediate checkpoints are stored as branches
(\texttt{step0}, then \texttt{chck\_1M} to \texttt{chck\_100M}), so the
training dynamics in Sections~\ref{sec:vpswap} and~\ref{sec:mechanism} can
be reproduced without retraining. The region features and the per-token
embedding tables that implement the seeding are released as datasets; the
latter would be needed for another model in order to reuse the intervention.

\section{Official BabyLM Evaluation}
\label{sec:official}

\paragraph{Measuring the effect.}
Every result below compares a vision-init model to its same-vocabulary
baseline; a \emph{delta} is vision-init minus baseline accuracy, in
points, and all zero-shot tasks are scored as minimal pairs under masked-language-model
pseudo-log-likelihood (mask each token in turn, sum the log-probability of
the original). Because single runs give no variance estimate, I look at
\emph{sign-consistency}: a task whose delta is positive in all nine
encoder$\times$vocabulary configurations is unlikely under a no-effect null
even when each delta is small. For the targeted analyses of
Section~\ref{sec:vpswap} I add three tools: (a) pair-level scoring (the two mirror items built from a noun pair are counted together, so a model's preference for one noun over the other cancels); (b) frequency reweighting (group accuracies compared after reweighting to a common corpus-frequency profile); and (c) \emph{McNemar's test} \citep{mcnemar1947}, which assesses
whether the two paired models differ significantly on the items where they
disagree. Following the diagnosis in Appendix~\ref{app:entity}, I also
check every minimal-pair task at initialization, where an unbiased probe
should be at chance.

\begin{figure}[t]
\centering
\includegraphics[width=\columnwidth]{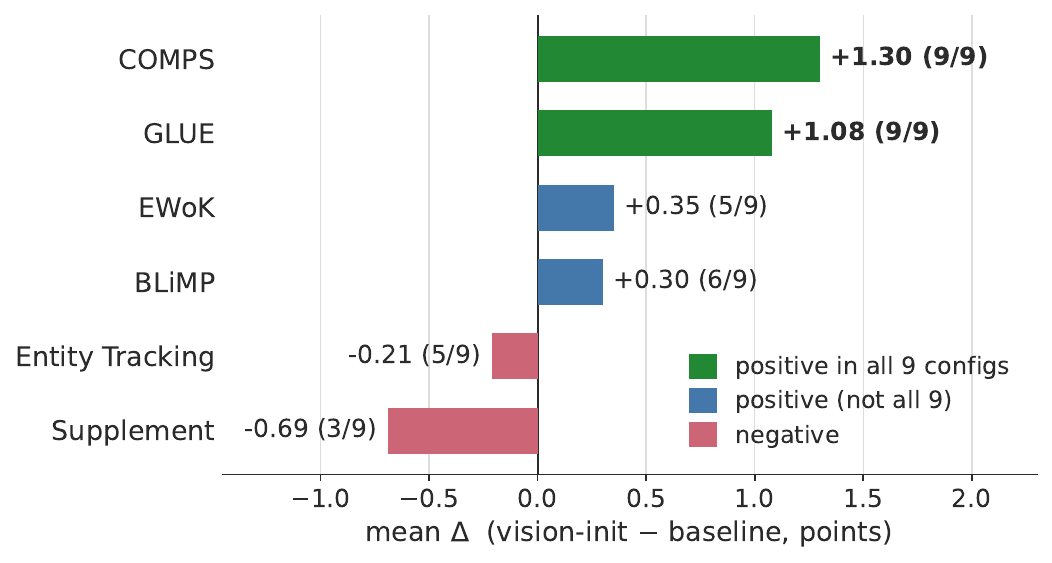}
\caption{Per-task deltas on the official BabyLM 2026 evaluation, averaged over the nine
encoder$\times$vocabulary configurations; the count is how many of the
nine are positive. Only object-property knowledge (\comps) and \glue are
positive in every configuration.}
\label{fig:official}
\end{figure}

\paragraph{An effect where expected.}
Figure~\ref{fig:official} gives the per-task deltas across all nine
encoder$\times$vocabulary configurations. Two tasks are
consistently positive: object-property knowledge (\comps, $+1.30$, 9/9)
and \glue ($+1.08$, 9/9). The rest are near zero and
inconsistent: \blimp and \ewok split roughly evenly
across configurations, and the entity-tracking and supplement tasks are
slightly negative. This is expected given the coverage analysis
(\S\ref{sec:coverage}): a vision prior should help with knowledge of what objects are like, and \comps{} is the only standard
zero-shot task that probes it directly.\footnote{The \glue effect is harder to attribute: \glue is evaluated by fine-tuning, and I decided not to investigate in detail how exactly my initialization gets exploited in the task-specific further training.} 

The three vision encoders behave the same: \comps{} and \glue are
positive in all nine configurations (three vocabularies $\times$ three encoders), so the effect does not depend on the choice of vision model. I
observe the same encoder-agnostic behaviour in the artifact of Appendix~\ref{app:entity}. For brevity I report the remaining analyses with a single representative encoder (SAM).

The \comps{} gain is expected: \comps{} pairs a
concept with a property and asks the model to prefer the true sequence;
its subconditions vary how the pair is presented. The gain is concentrated
in the conditions that test property knowledge itself (the base condition
$+1.37$, 9/9 and the harder novel-concept ``wugs'' condition $+3.65$, 9/9) and vanishes in the distractor conditions ($\approx0$),
where surface cues dominate. Vision-init helps the model know that, say, a banana is yellow.

\paragraph{The effect is stable.}
Because all configurations are single runs, I retrained the headline
pair (\sammod{} vs.\ \basemod) with three seeds and re-ran the full
zero-shot suite. \comps{} is the only zero-shot task whose sign is
stable across all three seeds ($+1.46/+1.05/+0.77$); all other zero-shot 
tasks change signs depending on the random seed (\blimp, \ewok, supplement, entity tracking). 

Zooming in on \ewok subtasks buttresses the same conclusion: gains concentrate in visual and quantitative content (\textit{number} $+6.53$, \textit{quantitative-properties} $+3.08$, \textit{material} $+2.15$) and disappear for social or material knowledge (Appendix~\ref{app:ewok}).

\paragraph{Leaderboard standing.}
I submitted my models to the official BabyLM 2026 leaderboard \citep{choshen2026babylm}, and they proved to be competitive with the field despite the intervention touching only a fraction of the vocabulary. The result should be read as a progression across my three models (all figures as of 19
July 2026). The text-only baseline \texttt{deberta-base-75k} scores 37.91
overall; adding visual seeding (\texttt{deberta-base-75k-sam}) raises this
to 39.63; and adding the synthetic-grounding extension (to be described in 
Section~\ref{sec:ext}) (\texttt{deberta-base-75k-sam\_ext-s1}) raises it
again to 40.62, which ranks 5th of the strict-small submissions ordered by overall average and sits
above the strongest official baselines (e.g.\ the Interaction baseline at
38.71). The same ordering holds, more sharply, on the Human-like Average:
the baseline scores 0.61, the seeded model 7.74, and the extended model
12.19 -- the highest of any entry on the board. I report the Human-like
figures with a caveat, since they are driven largely by the
age-of-acquisition component. 

\section{\vpswap: A Visual-Property Probe}
\label{sec:vpswap}
The official evaluation showed the effect only through a single zero-shot task
(\comps) and left its size and specificity unclear. To explore this effect further, I re-build \vpswap \citep{egobabyvlm}, a
minimal-pair probe of the knowledge a visual prior should bring in. In my version of the benchmark,  each item is annotated with the training frequency and seeded status of its words, allowing me to ask which words carry the effect.

\begin{figure*}[tbp]
  \centering
  \includegraphics[width=0.9\textwidth]{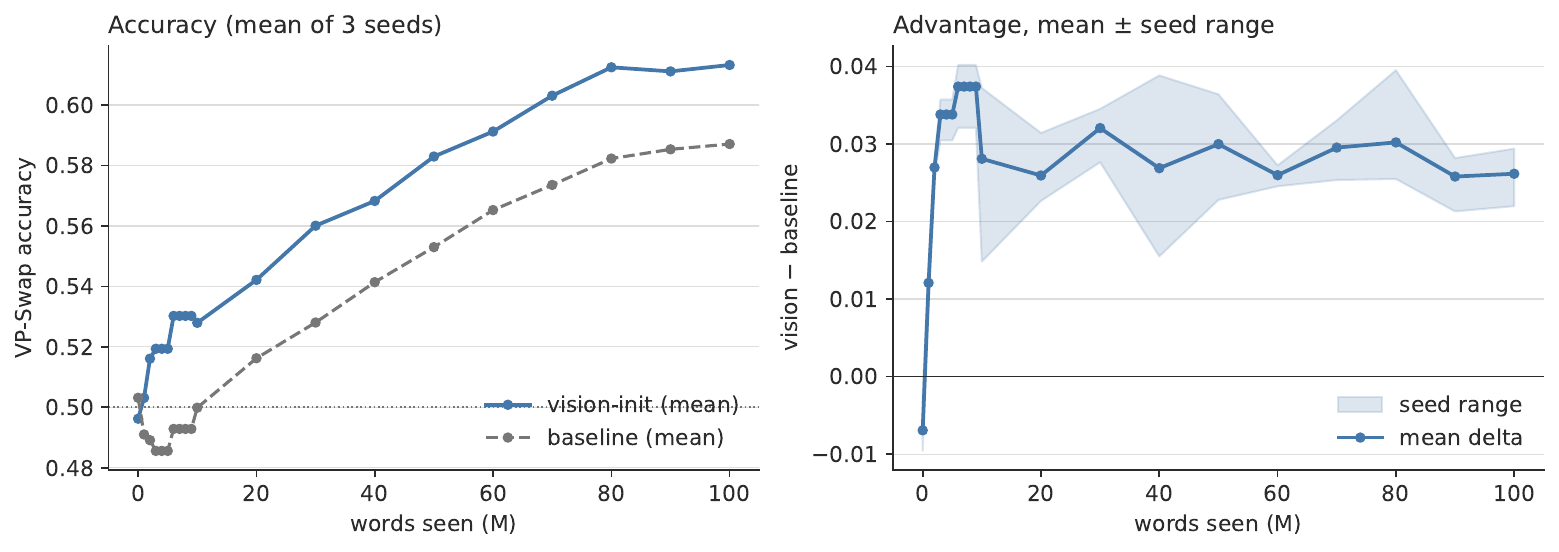}
  \caption{\vpswap{} accuracy over training, SAM-seeded (75k) vs.\ its
  baseline, three random seeds. The visually seeded model leads from $\sim$1M words to the
  end of training.}
  \label{fig:vpswap-traj}
\end{figure*}

\begin{figure*}[tbp]
  \centering
  \includegraphics[width=\textwidth]{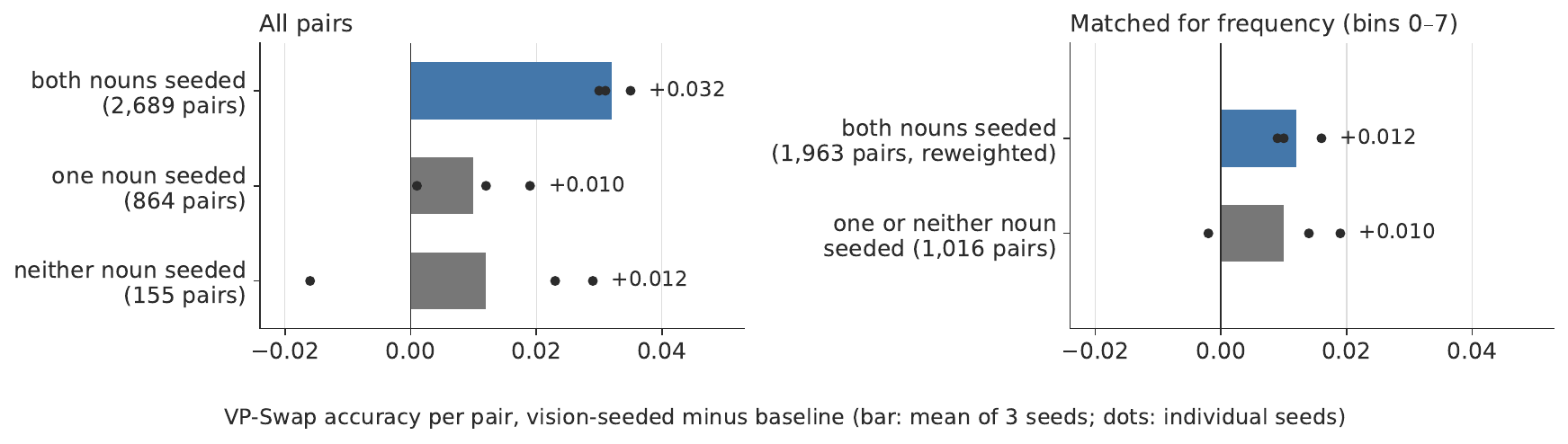}
  \caption{VP-Swap advantage by noun pair. Left: pairs by how many nouns were seeded. Right: after reweighting to a matched frequency profile, pairs with and without seeded nouns gain alike.}
  \label{fig:vpswap-2x2}
\end{figure*}

\subsection{Benchmark Construction}
Each VP-Swap item is a minimal pair in which a noun is
swapped for one whose typical properties are incompatible with the
property attributed to it, as in \textit{A cucumber is green} vs.\ \textit{A dune
is green}. The model is scored as correct when it assigns the higher
pseudo-log-likelihood to the compatible option. The benchmark covers four property types (color, material,
relative size, shape) across four syntactic frames (copular, attributive,
existential, relative). Each line contributes two items, since each of
the two sentences is also scored against the other's noun, for 7,416
items in total. Every item is generated from and keyed to my training corpus \corpus so that it records
each word's corpus frequency and whether that word was visually seeded.
Items are produced by an LLM generator and filtered by a separate LLM
judge; frequency bins follow the logarithmic edges of LongTail-Swap, VP-Swap's
sibling benchmark in the same suite, with all ten bins
populated per property (construction details and the full funnel are in
Appendix~\ref{app:vpswap}). At initialization, untrained checkpoints score
$0.49$--$0.51$, confirming the probe is unbiased. I release my version of
\vpswap, along with all code and models.

\subsection{Results}
\vpswap makes the initialization effect more visible and more explorable. Averaged over the
benchmark, the SAM-seeded model leads from very early in training -- by 1M words in all three seeds -- and keeps the advantage of roughly two to three points to the end (final delta $+2.2/+2.7/+2.9$ across seeds;
McNemar $z = 4.37/5.44/5.79$, all significant; Figure~\ref{fig:vpswap-traj}).
This is the same effect \comps{} detected, revealed by the dedicated probe.

\paragraph{Is the effect specific to seeded words?} Counted at the level of noun pairs (the two mirror items of a pair judged together, so that a preference for one noun cancels), pairs with two seeded nouns gain $+0.031/+0.030/+0.035$ across seeds, while pairs with one or no seeded noun gain about $+0.01$ (Figure~\ref{fig:vpswap-2x2}, left). But seeded nouns are the frequent nouns: the two highest frequency bins contain 726 both-seeded pairs and 3 pairs with an unseeded noun. Reweighted to a matched frequency profile, the two groups gain the same: $+0.012$ vs.\ $+0.010$ (Figure~\ref{fig:vpswap-2x2}, right). Word-level seeded status is also a poor proxy for what was seeded: 76\% of nominally unseeded nouns contain seeded subword tokens.

\paragraph{Item-level splits are misleading.} Split item by item, one seed shows $+0.051$ when the seeded noun is the correct one and $-0.049$ when it is the swapped-in one: the model prefers sentences with seeded nouns, whether correct or not. Per pair, this preference cancels ($+0.001$).

The gain holds across syntactic frames (attributive
$+0.023$, existential $+0.031$, relative $+0.027$) and is near zero in the
short copular frame ($-0.001$). It is largest for color
($+0.035$) and concentrated in mid- and high-frequency words; for the
rarest words both models sit near chance, consistent with the coverage
picture. Full per-property, per-frame, and per-bin tables are in
Appendix~\ref{app:tables}.

\section{Synthetic Grounding}
\label{sec:ext}
The evidence for the role of ostension so far is correlational. Moreover, \S\ref{sec:vpswap} showed it cannot even be located on the seeded words, because visual grounding and frequency correlate. An intervention would address both problems at once: ground words that were not grounded before and check whether the effect follows. The newly grounded words are the infrequent ones the confound obscures, so if grounding causes the effect, they should improve on \vpswap{} while already-seeded and still-unseeded words stay unchanged.

\paragraph{Extending grounding synthetically.}
For 1{,}986 concrete words with no image support, I generate scene descriptions with an LLM, render each as
images with a text-to-image model, detect the target object with an
open-vocabulary detector, and pool the same SAM features used for real
grounding inside the detected boxes; words the detector cannot find are
dropped. This grounds 1{,}155 previously unseeded words (adding 737 seeded
tokens), and I retrain the model from this extended initialization with
three random seeds (\extmod; pipeline details in Appendix~\ref{app:synth}). As a result, \vpswap items fall into three groups: those whose noun was
\emph{really} seeded (6{,}242 items, a control that should not change),
those whose noun is now \emph{synthetically} seeded (835 items, the
treated group), and those still unseeded (339 items, a second control).

\begin{figure}[t]
  \centering
  \includegraphics[width=\columnwidth]{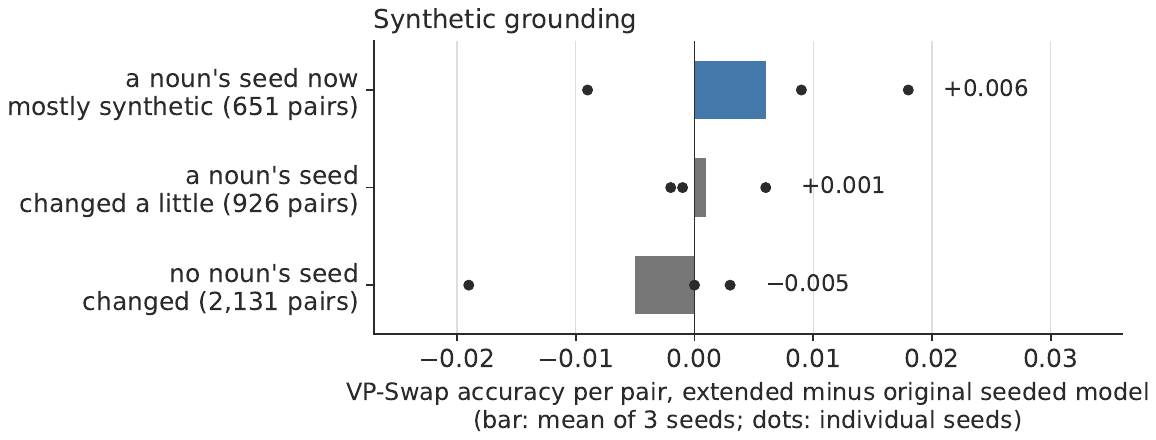}
  \caption{Pair-level \vpswap{} advantage over the text-only baseline by
  grounding group, for the seeded model (\sammod) and the synthetic
  extension (\extmod). The extension adds nothing in any group.}
  \label{fig:ext}
\end{figure}

\paragraph{The effect does not follow the grounding.} At the item level the treated group appears to improve in all three seeds, but the improvement does not survive pair-level counting (Figure~\ref{fig:ext}): on treated pairs the extension is not distinguishable from the SAM model, and on the whole benchmark \extmod{} $-$ \sammod{} is $+0.5/0.0/-1.0$ points (\comps: $+0.11/+0.11/-0.37$). The extension preserves the seeding gain over the text-only baseline but adds nothing to it. The reason is visible in the seeds themselves. The target words were packed into shared scene descriptions (6.2 targets per generated image), so a word's seed averages about four regions from a single scene, and the resulting vectors are nearly identical: the mean cosine between two synthetic seeds is $0.71$, against $0.002$ between two real seeds (Appendix~\ref{app:synth}). \comps{} remains positive over the baseline in every seed ($+1.57/+1.16/+0.40$), so the extension does not disturb the original effect.

\section{Mechanism: A Prior with a Lasting Imprint}
\label{sec:mechanism}
How does an intervention applied only at initialization still shape a model
after 100M words of training? 
 I track each seeded embedding's cosine
similarity to its initial value over training: it falls from $1.00$ to
about $0.15$ by 100M words -- the embeddings end up nearly
orthogonal to where they began. This means that whatever is driving the effect of the visual initialization, it is not
the survival of the original visual vectors.
 If the benefit came from staying close to the visual initialization, words
that drifted less should benefit more. That is not the case: per-word drift is essentially uncorrelated with the change in that word's \vpswap advantage
($r=-0.017$ over the 816 seeded words with at least three items).

\paragraph{The effect is relational.} While the absolute positions of the seeded embeddings move almost entirely, the \emph{relative} geometry among
them is partly preserved. Measured by
representational similarity analysis (RSA) \citep{kriegeskorte2008rsa}, the
similarity structure of the seeded embeddings still correlates with that
of their visual anchors at 100M words (RSA $0.31$). The natural comparison
is the baseline model: computed over the very same words, its embeddings
show almost no such structure (RSA $0.10$), confirming that a text-only
model does not recover this visual geometry on its own -- it is present only
because those words were visually initialized. The correlation is stable
over the second half of training. The visual prior thus leaves a lasting
imprint on how seeded words are arranged relative to one another, even if each word's individual position is overwritten by distributional learning.

\section{The Limits of Ostension}
\label{sec:scope}
\begin{figure*}[t]
  \centering
  \includegraphics[width=.94\textwidth]{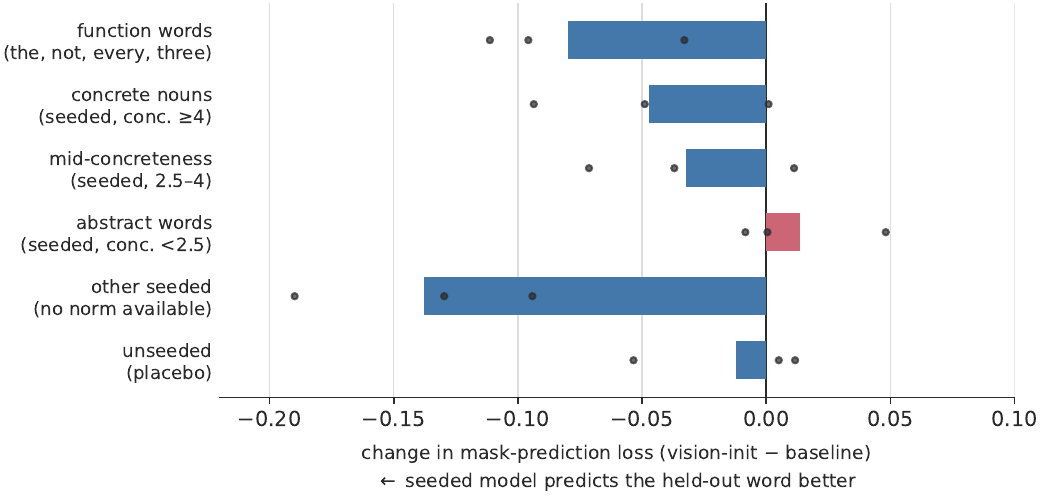}
  \caption{Change in held-out mask-prediction loss when words are
  vision-seeded (vision-init $-$ baseline), by token class, under
  identical masking. Bars are the mean over three seeds, dots the
  individual seeds. Seeded classes become easier to predict, the never-seeded placebo class at
  $\approx 0$.}
  \label{fig:scope}
\end{figure*}

So far the story has mostly been about concrete nouns, as suggested by the  
coverage analysis. But grounding data does not
only contain concrete nouns. Captions also contain function words and abstract vocabulary -- \textit{the},
\textit{every}, \textit{three} -- and these words, too,
received visual initializations, from the average of the many scenes they
appear in. Do those seeds matter? I show that ostension is important even for the meaning of these words, but this is an effect that existing benchmarks do not register.

\paragraph{Visual initialization for abstract vocabulary?}
One might expect a function word appearing in a lot of image contexts to receive a washed-out average vector. It is partly the case: at matched image support, function-word seeds are weaker than concrete controls (centered magnitude 5.1 vs. 11.1 for words with over 1000 supporting regions; both sets are listed in Appendix~\ref{app:wordlists}). They are, however, far from washed out to zero. Moreover, the model retains the visual information to the end of training. The relational structure among function-word
embeddings still reflects the visual anchor at 100M words (RSA $0.45$;
the baseline model, computed over the same words, shows none, RSA
$\approx 0$), and their absolute similarity to the anchor is if anything
\emph{higher} than for concrete controls ($0.44$ vs.\ $0.34$). 

\paragraph{Used by the objective, not rewarded by benchmarks.}
However, on the benchmarks, these retained seeds do nothing. Across three
seeds, \blimp phenomena whose minimal pairs differ in an \emph{abstract}
word average $-0.8$ points (vision $-$ baseline), while phenomena differing
in a \emph{concrete} word average $+0.8$; the correlation between a
phenomenon's delta and the concreteness of its varying words is $r=+0.20$
over the 54 fast-subset phenomena.
The structural competence these benchmarks test doesn't seem to be  something a visual prior for abstract vocabulary can improve. But this is a
fact about the benchmarks, not about whether the seeds are useful for language learning:
measured directly on the pretraining objective, they are useful.
Figure~\ref{fig:scope} reports held-out masked-LM loss by token class
under identical masking: the vision-initialized model predicts masked function words better than the baseline in all three seeds (a mean
cross-entropy reduction of $0.080$ nats), with never-seeded tokens at $\approx 0$. The retained visual
structure is useful for the training objective.

There is one suggestive exception. The only abstract domain with consistent vision-related improvement is \ewok \textit{number} ($+5.8$). Numerals are quite well visually seeded, and quantity is abstract content that nonetheless is visually manifest, so a visual prior  plausibly carries information about it. The samples are small and I treat this as speculation.

Summing up, the scope of ostension in language learning is broader than one might think, and broader than what we are currently able to measure. The visual seeds are strong even for functional words that are usually thought of as lacking visual aspects of meaning; moreover, the model's training objective finds these visual properties useful. Existing evaluations mostly test abstract structural competence and are
blind to these effects. What evaluation would register them is an open question.
 
\section{Discussion}
I showed that the effect of visual grounding is real, but barely visible in the aggregate BabyLM evaluation. I would like to draw the field's attention to this and potential other cases where intervention and evaluation are misaligned in scope. 

I demonstrate the effect with a targeted, metadata-rich probe informed by training dynamics -- items that are annotated with training frequency and seeding status, checked at chance at
initialization. I suggest such
probes, rather than averages alone, as an instrument
for measuring interventions that touch a specific and predictable part of
what a model knows. The null result from EgoBabyVLM's text-encoder
ablation \citep{egobabyvlm}, from a different intervention,
reinforces the point: whether grounding ``helps'' isn't enough without clarifying which parts of vocabulary it helps and under what measurements.

Mechanism analysis (\S\ref{sec:mechanism}) has consequences for deciding where to invest effort to magnify the effects of injection. Since the visual information survives as relational structure
rather than as a retrievable anchor, interventions
that periodically re-inject visual embeddings during training are not likely to work; richer grounding at initialization is more promising. And since the visual seeds for function words are strong, retained, and
used by the objective but not rewarded by any benchmark (\S\ref{sec:scope}), an
obvious next step is evaluation designed to test what a visual prior could actually contribute to, see the suggestive \ewok \textit{number} effect above. 
 
\section*{Limitations}
Most configurations are single runs; three model types were trained with three seeds each: the 75k baseline, its SAM-seeded counterpart, and the synthetic extension. My corpus is not the official 2026 strict-small distribution, which limits
direct comparison to submissions trained on the standard data. \vpswap{} is generated and filtered by LLMs and  inherits their
notion of typical object properties.  Finally, several analyses have reduced
scope: concreteness-band results cover only the $\sim$11k of 21k seeded
tokens with Brysbaert norms, the curated function-word probe is 64 words,
and the \blimp concreteness gradient uses the fast-evaluation subset. All
results are for a single masked-LM architecture. Finally, two limitations
of the \vpswap{} attribution analysis: seeded status is confounded with
corpus frequency, so the benchmark cannot attribute the gain to the seeded
words; and seeded status is annotated at the word level while seeding
applies to subword tokens -- 76\% of nominally unseeded nouns contain
seeded tokens.
 
\section*{Ethics Statement}
This work carries low ethical risk. The models are trained from
scratch on a public developmentally motivated corpus, and are not intended for deployment. Two components of my method rely on generative models. My version of the \vpswap benchmark is constructed with an LLM
generator and an LLM judge; the resulting items  reflect those
models' notions of typical object properties, which may encode cultural or
distributional biases, and the released benchmark should be read as a
diagnostic probe rather than a source of ground-truth world knowledge. My
synthetic-grounding extension (\S\ref{sec:ext}) generates images with a
text-to-image model and localizes objects with an open-vocabulary
detector; these components can hallucinate or mis-detect, and the pipeline
inherits whatever biases they carry. Because both pipelines feed only
model initialization and a minimal-pair evaluation -- never user-facing output -- I judge these risks as contained, but I flag them so that downstream users of the released artifacts can account for them. All
datasets and models used are publicly available and used consistently with their intended research purpose. 

\section*{Acknowledgments}
I thank Gabriele Sarti for the original idea behind the image-derived
embeddings and for pointing me to the VP-Swap benchmark; Rick Nouwen for
connecting visual initialization to Augustine's account of word learning;
and Ece Takmaz for suggesting concreteness bins as a lens on the benchmarks,
and for much other discussion. I am grateful to Lukas Edman and Jakub
Dotla\v{c}il for helpful discussions, and to the anonymous reviewers for
their comments. This work used the Dutch national e-infrastructure with the support of the
SURF Cooperative using grant no.\ EINF-18223.

\bibliography{references}
 
\appendix

\section{Training Hyperparameters}
\label{app:hparams}
All models use the same recipe (Table~\ref{tab:hparams}), following the
configuration of \citet{edman2024babylm}. Only the input-embedding
initialization differs between a vision-init model and its baseline; every
other hyperparameter, the data order, and the random seed are held fixed,
so the two model pairs match as close as possible.

\begin{table}[h]
\centering
\footnotesize
\setlength{\tabcolsep}{4pt}
\begin{tabular}{@{}ll@{}}
\toprule
Hyperparameter & Value \\
\midrule
Architecture & \deberta-base (MLM) \\
Hidden size / layers / heads & 768 / 12 / 12 \\
Intermediate size & 3072 \\
Vocabulary (BPE) & 50k / 75k / 100k \\
Learning rate & $2\times10^{-4}$ \\
LR schedule & cosine \\
Warmup & 4000 steps \\
Weight decay & 0.01 \\
Optimizer & AdamW ($\beta_1{=}0.9$, $\beta_2{=}0.95$) \\
Effective batch size & 256 (accum.\ 4) \\
Epochs & 10 \\
Context-length warmup & 64$\to$128 (from ep.\ 5) \\
Max sequence length & 512 \\
\bottomrule
\end{tabular}
\caption{Training hyperparameters, shared across all baselines and
vision-init models.}
\label{tab:hparams}
\end{table}

\section{\ewok Subtask Deltas}
\label{app:ewok}
Table~\ref{tab:ewok} gives the per-subtask EWoK deltas (vision-init minus
baseline, averaged over the nine encoder$\times$vocabulary configurations,
\texttt{best} revision). Subtasks
whose content is visual or quantitative (\textit{number},
\textit{quantitative-properties}) show the largest and
most consistent gains, while socially or spatially framed subtasks are
flat or negative.

\begin{table}[h]
\centering
\small
\begin{tabular}{lrr}
\toprule
Subtask & Mean $\Delta$ & Positive \\
\midrule
number & $+6.53$ & 6/9 \\
active-passive & $+4.81$ & 7/9 \\
quantitative-properties & $+3.08$ & 8/9 \\
material-dynamics & $+2.51$ & 6/9 \\
material & $+2.15$ & 6/9 \\
direct & $+1.02$ & 8/9 \\
negation & $+0.76$ & 5/9 \\
concept swap & $+0.59$ & 6/9 \\
physical-dynamics & $+0.47$ & 4/9 \\
physical-relations & $+0.45$ & 4/9 \\
social-relations & $+0.32$ & 5/9 \\
antonym & $+0.26$ & 7/9 \\
agent-properties & $+0.14$ & 5/9 \\
physical-interactions & $+0.10$ & 5/9 \\
indirect & $+0.01$ & 5/9 \\
material-properties & $+0.00$ & 4/9 \\
other & $-0.16$ & 3/9 \\
social-interactions & $-0.76$ & 3/9 \\
spatial-relations & $-1.02$ & 1/9 \\
social-properties & $-1.42$ & 2/9 \\
game & $-4.44$ & 0/9 \\
\bottomrule
\end{tabular}
\caption{\ewok subtask deltas (vision-init $-$ baseline), mean over 9
configurations. ``Positive'' counts configurations with a positive delta.}
\label{tab:ewok}
\end{table}

\section{\vpswap{} Construction Details}
\label{app:vpswap}
\vpswap contains 7{,}416 minimal-pair items across four properties
(color, material, relative size, shape), built from \corpus following the
Visual-Property Swap protocol of \citet{egobabyvlm} with the deviations
noted below. Each pair-file line encodes two sentences and the character
offsets of the swapped nouns; metadata records each noun's corpus
frequency, seeded status, property, and syntactic frame.

\paragraph{Generation funnel.} Starting from corpus nouns with Brysbaert
norms, items pass a sequence of automatic gates
(Table~\ref{tab:funnel}). Sentences are generated by
\texttt{claude-sonnet-4-6} (temperature 0.7); a concreteness/inanimacy
gate and a 4$\times$ A/B plausibility judge use \texttt{claude-haiku-4-5}
(temperature 0), and only items the judge accepts on all four trials
survive. Frequency bins follow the logarithmic edges
$[1, 2, 4, \ldots, 512]$ used by LongTail-Swap
\citep{egobabyvlm}, the frequency-controlled sibling of VP-Swap in the
same benchmark suite; all ten bins are populated for every property.

\begin{table}[h]
\centering
\footnotesize
\begin{tabular}{lr}
\toprule
Stage & Count \\
\midrule
Corpus nouns w/ Brysbaert norms & 11{,}921 \\
Concreteness gate ($\geq 4.0$, noun) & 5{,}006 \\
Inanimacy gate (LLM) & 1{,}669 \\
\quad seeded / unseeded & 1{,}306 / 363 \\
Sampled pairs per property & 2{,}000 \\
Well-formed generations & 1{,}739--1{,}864 \\
Attribution gate & 1{,}160--1{,}259 \\
4$\times$ A/B judge (all-correct) & 878--992 \\
Final (per property) & 865/964/944/935 \\
\bottomrule
\end{tabular}
\caption{\vpswap{} construction funnel. Final counts are per property
(color / material / relative-size / shape); each pair yields two
minimal-pair items, for 7{,}416 total.}
\label{tab:funnel}
\end{table}

\paragraph{Syntactic frames.} Sentences rotate over four frames -- copular (\textit{A femur is white}), attributive (\textit{the white femur}), existential,
and relative-clause -- so the effect can be checked for robustness to sentence form; the judge rejects copular items at a somewhat higher rate.

\paragraph{Deviations from the upstream protocol.} I follow the
Visual-Property Swap task of \citet{egobabyvlm}
(the \texttt{visual\_property\_swap} benchmark) with several deliberate
changes. All items are generated from and frequency-binned against my own
training corpus \corpus rather than a fixed external word list, so that
difficulty bins reflect the frequencies this model actually saw. The
generator and judge models are my own (\texttt{claude-sonnet-4-6} and
\texttt{claude-haiku-4-5}) with the four-trial A/B acceptance gate
described above. Item selection is stricter: upstream uses an LLM
yes/no physical-object judgment, whereas I gate on Brysbaert
concreteness and part of speech, then add the inanimacy and attribution
gates of the funnel above. Sentences rotate over fixed syntactic frames
rather than being generated freely, and swap indices are validated at word
boundaries, so that a substring match cannot corrupt a swap (\emph{tub}
inside \emph{bathtub}). Most important for my analysis, every item is
annotated with the seeded status of both of its nouns, which is what makes the difference-in-differences and placebo
analyses of Section~\ref{sec:vpswap} possible. Finally, when a noun is substituted I adjust a preceding \emph{a} or
\emph{an} to agree with the incoming noun, so that the distractor never
differs from the original in grammaticality. Without this, swapping a
vowel-initial noun under \emph{a} yields an ungrammatical distractor
that a model can reject on agreement alone. The correct article is taken
from the generated sentences themselves where available, since they are a
better authority than a spelling rule (\emph{a unicycle}, \emph{a uterus}),
falling back to the vowel rule otherwise.

\section{Synthetic Grounding Pipeline Details}
\label{app:synth}
To isolate whether it's grounding that causes the effect
(\S\ref{sec:ext}), I extend coverage to previously unseeded concrete
words. From 1{,}986 concrete zero-support words I generate short scene
descriptions with \texttt{claude-sonnet-4-6} (temperature 0.8), prompting
it to place as many target words as fit naturally into a single scene.
Each description is rendered into three images with SDXL-Turbo
\citep{sauer2024sdxlturbo} (\texttt{stabilityai/sdxl-turbo}, 2 inference
steps, guidance scale 0). I then localize each target word with
open-vocabulary detection (\citealt{minderer2023owlv2}, \texttt{owlv2-base-patch16-ensemble}, score threshold 0.25). Words that
the detector never finds in any of the images are dropped, so a word enters the extended seed set only if it can actually be pictured and found. SAM features are pooled inside the detected boxes through the same Stage-1 extraction code used for real grounding. This yields 1{,}155
newly grounded words ($+737$ seeded tokens, $21{,}134 \to 21{,}871$), from
which I train \extmod{} with three seeds.

Because a noun's ``unseeded'' status in \vpswap{} is defined against the original seed set, extension induces a three-way item split: real seeded / synthetically seeded / still-unseeded (see Section~\ref{sec:ext}). The still-unseeded group
(339 items) serves as a within-benchmark control, and the real-seeded group (6{,}242 items) should be untouched by extension.

\begin{figure*}[t]
  \centering
  \includegraphics[width=\textwidth]{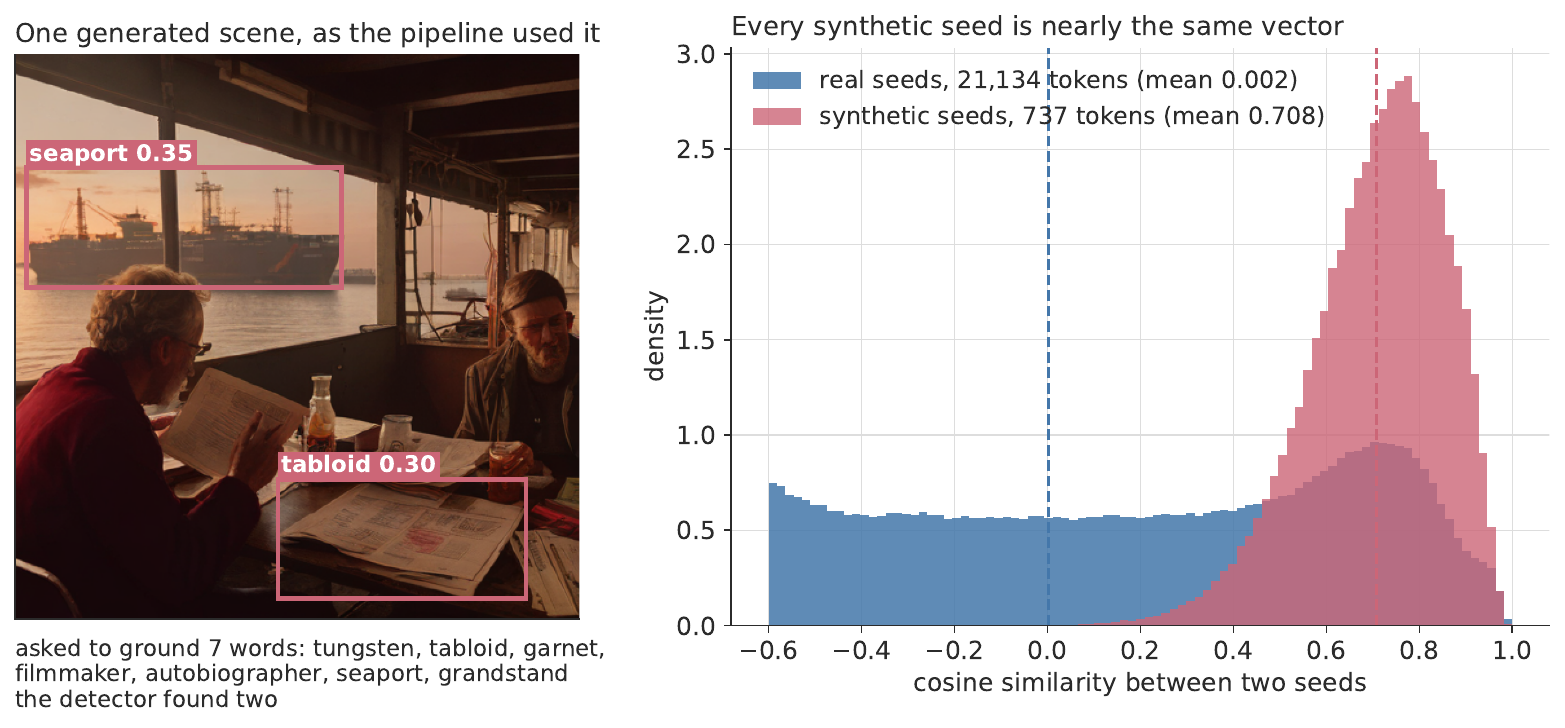}
  \caption{Left: one generated scene with the regions the detector found;
  the scene description was written to contain seven target words. Right:
  cosine similarity between pairs of seeds in the initialization table:
  real seeds spread out, synthetic seeds nearly coincide.}
  \label{fig:synthwhy}
\end{figure*}

\paragraph{Why the synthetic seeds collapse.} The 1,986 targets were placed into 1,054 shared scene descriptions (6.2 per image, up to 10), each rendered three times; a word's seed averages a median of 3 regions, all from one scene. The mean pairwise cosine among the 737 synthetic-only token seeds is $0.71$, against $0.002$ among real seeds (Figure~\ref{fig:synthwhy}). Same-scene words are barely more similar (0.76) than different-scene words (0.71), so the shared component is the generator's rendering style, not scene content; detection is not the weak point (1.7 of $\sim$6 targets found per image, median score 0.36, found boxes correct).

\section{Full Results Tables}
\label{app:tables}
This appendix collects the breakdowns referenced from the main text. The
per-task official deltas and the \ewok subtask deltas appear in
Figure~\ref{fig:official} and Appendix~\ref{app:ewok}; the coverage
figures are in the main text. Table~\ref{tab:comps-sub} gives the
\comps{} subtask breakdown, and Table~\ref{tab:vpswap-break} the \vpswap{}
results by property and by syntactic frame.

\begin{table}[h]
\centering
\footnotesize
\setlength{\tabcolsep}{5pt}
\begin{tabular}{@{}lrr@{}}
\toprule
\comps subtask & Mean $\Delta$ & Positive \\
\midrule
base (property knowledge) & $+1.37$ & 9/9 \\
wugs (novel concept) & $+3.65$ & 9/9 \\
wugs-dist-before & $+0.15$ & 4/9 \\
wugs-dist-in-between & $+0.04$ & 4/9 \\
\bottomrule
\end{tabular}
\caption{\comps{} subtask deltas (vision-init $-$ baseline), mean over 9
configurations. The gain is in the property-knowledge conditions and
vanishes when distractors are added.}
\label{tab:comps-sub}
\end{table}

\begin{table}[h]
\centering
\footnotesize
\setlength{\tabcolsep}{4pt}
\begin{tabular}{@{}lrrr@{}}
\toprule
Group & $n$ & vision & $\Delta$ \\
\midrule
\multicolumn{4}{@{}l}{\emph{By property}}\\
color & 1730 & 0.639 & $+0.035$ \\
material & 1928 & 0.646 & $+0.015$ \\
relative size & 1888 & 0.575 & $+0.015$ \\
shape & 1870 & 0.594 & $+0.024$ \\
\midrule
\multicolumn{4}{@{}l}{\emph{By syntactic frame}}\\
attributive & 1890 & 0.658 & $+0.023$ \\
copular & 1244 & 0.469 & $-0.001$ \\
existential & 1680 & 0.652 & $+0.031$ \\
relative & 2602 & 0.624 & $+0.027$ \\
\bottomrule
\end{tabular}
\caption{\vpswap{} accuracy (SAM-seeded 75k) and delta vs.\ baseline, by
property and by syntactic frame, seed \texttt{s1}. The effect is largest
for color and holds in every frame but the short copular one, where both
models sit slightly below chance.}
\label{tab:vpswap-break}
\end{table}

\section{Grounding Scope: Additional Analyses}
\label{app:scope}
Supporting detail for Section~\ref{sec:scope}. Table~\ref{tab:rsa} reports
representational-similarity \citep{kriegeskorte2008rsa} between seeded
embeddings and their visual anchors at the end of training (100M words),
by concreteness band: the relational structure of the visual seeds
survives training across all bands, and is retained most strongly
for function words, even though, as the main text shows, no benchmark rewards it. The full
function-word and concrete-control sets are listed in
Appendix~\ref{app:wordlists}.

\begin{table}[h]
\centering
\footnotesize
\setlength{\tabcolsep}{4pt}
\begin{tabular}{@{}lrrr@{}}
\toprule
Group & $n$ & vision-init & baseline \\
\midrule
band 1--2 (abstract) & 731 & 0.188 & $-0.011$ \\
band 2--3 & 2186 & 0.189 & 0.001 \\
band 3--4 & 3123 & 0.131 & $-0.004$ \\
band 4--5 (concrete) & 5006 & 0.180 & 0.034 \\
function words & 64 & 0.446 & $-0.055$ \\
concrete controls & 30 & 0.741 & 0.250 \\
\bottomrule
\end{tabular}
\caption{RSA between seeded embeddings and their visual anchors at 100M
words, by concreteness band, with the text-only baseline floor. Visual
structure is retained across all bands (baseline $\approx 0$), and among
the abstract vocabulary is strongest for function words.}
\label{tab:rsa}
\end{table}

The values in Table~\ref{tab:rsa} are computed on raw cosines. Centring each space leaves the function-word and abstract-band values intact (function words $0.68$ vs.\ a $0.16$ baseline; band 1--2 at $0.22$) but lowers large concrete-noun sets substantially (1{,}011 seeded \vpswap{} nouns: $0.31 \to 0.09$, baseline $0.10 \to 0.02$).

Table~\ref{tab:mlmloss} gives the held-out MLM loss by token class (vision-init $-$ baseline, three seeds and their mean). The objective is easier for seeded classes -- most of all for function words ($-0.080$ nats, 3/3 seeds) -- while the never-seeded class sits at the $\approx 0$ placebo.

\begin{table}[h]
\centering
\footnotesize
\setlength{\tabcolsep}{3pt}
\begin{tabular}{@{}lrrrrr@{}}
\toprule
Class & $n$ & s1 & s2 & s3 & mean \\
\midrule
function & 1385 & $-.096$ & $-.033$ & $-.111$ & $-.080$ \\
abstract ($<2.5$) & 1445 & $-.008$ & $+.048$ & $+.001$ & $+.013$ \\
mid (2.5--4) & 1327 & $+.011$ & $-.037$ & $-.071$ & $-.032$ \\
concrete ($\geq 4$) & 589 & $+.001$ & $-.094$ & $-.049$ & $-.047$ \\
seeded-other & 2446 & $-.130$ & $-.094$ & $-.190$ & $-.138$ \\
unseeded (placebo) & 1140 & $+.005$ & $+.012$ & $-.053$ & $-.012$ \\
\bottomrule
\end{tabular}
\caption{Held-out MLM loss by token class (vision-init $-$ baseline, in
nats; negative $=$ vision-init predicts the masked token better).
Identical masks across models. Seeded classes improve; the unseeded class
is $\approx 0$.}
\label{tab:mlmloss}
\end{table}

The \blimp concreteness gradient (correlation of a phenomenon's vision-init
delta with the concreteness of its varying words) is $r=+0.20$ over the 54
fast-subset phenomena. 

\section{Word Lists for the Scope Analysis}
\label{app:wordlists}
The scope analysis of Section~\ref{sec:scope} and
Appendix~\ref{app:scope} uses two curated word sets. The \emph{function-word} set (64 items) collects negation, quantifiers, determiners, connectives, numerals, and a few high-frequency deictics,  words standardly treated as lacking perceptual meaning:

\begin{quote}\small\ttfamily
not, no, never, nothing, nobody, none, neither, nor, every, each, all,
some, any, few, many, most, several, both, either, or, and, but, if,
unless, because, although, whether, than, as, only, even, also, too, very,
quite, must, might, may, could, should, would, will, can, one, two, three,
four, five, six, seven, eight, nine, ten, first, second, third, the, a,
an, this, that, these, those, there, it
\end{quote}

The \emph{concrete-control} set (30 items) is a set of high-concreteness,
picturable common nouns used as the comparison group:

\begin{quote}\small\ttfamily
dog, cat, table, chair, apple, banana, car, tree, house, ball, cup, door,
book, shoe, water, bread, bird, fish, horse, bed, spoon, window, bottle,
hat, box, stone, flower, truck, boat, chicken
\end{quote}

\section{The Entity-Tracking Scoring Artifact}
\label{app:entity}
During checkpoint-dynamics evaluation I initially observed a vision-init effect on the BabyLM entity-tracking task: a $+17$-point
spike at $\sim$step 1000 (50k vocabulary, all three encoders). Closer
analysis showed this to be a scoring artifact rather than entity-tracking
competence, and I exclude entity tracking from interpretation throughout.

\begin{itemize}
\item \textbf{Untrained MLMs score $\sim$42\%} on this task -- far above the
nominal chance level -- because the pseudo-log-likelihood scoring interacts with the answer options' length and frequency priors.
\item As training proceeds this inflated baseline \emph{collapses};
vision-init models merely delay the collapse by $\sim$1000 steps rather
than building competence.
\item \textbf{Inverted subtask difficulty} confirms the diagnosis:
0-operation items score \emph{lowest} and 5-operation items \emph{highest}
at every checkpoint, including the final one, which is the opposite of the true
difficulty ordering.
\item The spike is 50k-only; at 75k and 100k the corresponding peaks are
just $+1.0$ and $+1.7$.
\end{itemize}

I conclude that entity-tracking scores of MLM-scored models should be treated as an artifact. An untrained-model check at step 0 is a cheap, effective diagnostic that would have caught this immediately; I
now run it for every minimal-pair-style evaluation (e.g.\ \vpswap{} sits at $0.49$--$0.51$ at initialization, which is chance level).

\end{document}